\documentclass[runningheads]{llncs}
\usepackage{graphicx}
\usepackage{siunitx}
\usepackage{pbox}
\usepackage{array}
\usepackage{tabularx}
\usepackage{multirow}
\usepackage{rotating, booktabs}
\usepackage{csquotes}
\usepackage{amsfonts}
\usepackage{amsmath}
\usepackage{xcolor}
\usepackage{hyperref}
\usepackage{url}

\begin{document}
%
%\title{Contribution Title\thanks{Supported by organization x.}}
\title{Automated Scoring for Reading Comprehension via In-context BERT Tuning}
\author{Nigel Fernandez\inst{1} \and Aritra Ghosh\inst{1} \and Naiming Liu\inst{2} \and 
Zichao Wang\inst{2} \and Beno\^{\i}t Choffin \and Richard Baraniuk\inst{2} \and Andrew Lan\inst{1}}

\authorrunning{N. Fernandez et al.}
\institute{$^1$University of Massachusetts Amherst, $^2$Rice University \\
Contact Email: \email{andrewlan@cs.umass.edu}}
\maketitle              % typeset the header of the contribution
\begin{abstract}
Automated scoring of open-ended student responses has the potential to significantly reduce human grader effort. Recent advances in automated scoring often leverage textual representations based on pre-trained language models such as BERT and GPT as input to scoring models. Most existing approaches train a separate model for each item/question, which is suitable for scenarios such as essay scoring where items can be quite different from one another. However, these approaches have two limitations: 1) they fail to leverage item linkage for scenarios such as reading comprehension where multiple items may share a reading passage; 2) they are not scalable since storing one model per item becomes difficult when models have a large number of parameters. 
In this paper, we report our (grand prize-winning) solution to the National Assessment of Education Progress (NAEP) automated scoring challenge for reading comprehension. Our approach, in-context BERT fine-tuning, produces a single shared scoring model for all items with a carefully-designed input structure to provide contextual information on each item. We demonstrate the effectiveness of our approach via local evaluations using the training dataset provided by the challenge. We also discuss the biases, common error types, and limitations of our approach. 

\end{abstract}

%
% SECTION %
%
\section{Introduction}

Automated scoring (AS) refers to the problem of using algorithms to automatically score student responses to open-ended items. AS approaches have the potential to significantly reduce the amount of human grading effort and scale well to an ever-increasing student population. AS has been studied in many different contexts, especially essays, known as automated essay scoring (AES)~\cite{asap,aes}, but also in other contexts such as automatic short answer grading (ASAG)~\cite{irtasag,chiasag} for domains including humanities, social sciences~\cite{ibmin}, and math~\cite{sami}. 

Existing approaches for AS include feature engineering, i.e., developing features that summarize the length, syntax~\cite{erater,iea,aes}, cohesion~\cite{cohmetrix}, relevance to the item~\cite{relevance}, and semantics~\cite{semantic} of student responses combined with machine learning-based classifiers~\cite{attali2006automated,danielle} to predict the score. These approaches have excellent interpretability but require human expertise in feature development. Recent approaches leverage advances in deep learning to produce better textual representations of student responses, alleviating the need to rely on human-engineered features. Examples include combining neural networks with human-engineered features for score prediction~\cite{uto2020neural} and especially approaches that fine-tune neural language models (LMs) such as BERT~\cite{devlin2018bert} and GPT-2~\cite{radford2019gpt2} on the downstream AS task~\cite{cambium,mayfield2020should,taghipour2016neural,yang2020enhancing}. Despite concerns on the vulnerability against gaming input~\cite{cambium} and fairness aspects~\cite{diane} of these approaches, they have performed well in terms of scoring accuracy on publicly available benchmark datasets such as the automated student assessment prize (ASAP) for essay scoring~\cite{asap}. 

However, existing AS approaches are limited in many ways, including a significant one we address in this paper: They almost always train a separate AS model for each item. This approach is acceptable in contexts such as AES since the items (essay prompts) are likely not highly related to each other. However, in other contexts such as reading comprehension, multiple items may share the same background passage;  Training a separate model for each item fails to leverage this shared background information. More importantly, this approach results in a separate model for each item; for LM-based models that have millions of parameters, this approach creates a significant model storage problem.

% SUBSECTION %
\subsubsection{Contributions}
In this paper, we detail a novel AS approach for reading comprehension items via in-context fine-tuning of LMs, which is our (grand prize-winning) solution to the item-specific part of the National Assessment of Educational Progress (NAEP) automated scoring challenge.\footnote{Ran by the US Dept.\ of Education: \url{https://github.com/NAEP-AS-Challenge/info}} We make our implementation publicly available\footnote{Our code can be found at: \url{https://github.com/ni9elf/automated-scoring}} to facilitate future work on automated scoring. Our contributions are listed below:

\begin{itemize}
\item We develop an AS approach based on multi-task and meta-learning ideas, which fine-tunes an LM with a carefully designed input format that captures the context of each individual item. Our approach effectively leverages potential linkage across all items. 

\item We demonstrate the effectiveness of our AS approach through experimental evaluation on the NAEP challenge training dataset. Our approach outperforms existing LM-based AS approaches and significantly outperforms other non-LM-based baselines. 

\item We qualitatively analyze the cause of scoring errors made by our approach. We also show that our AS approach exhibits modest bias towards different student demographic groups. We outline a series of avenues for future work to improve the applicability of our approach in real-world settings. 

\end{itemize}

%
% SECTION %
%
\section{Methodology}

In this section, we first introduce the problem setup for the NAEP AS challenge (Section~\ref{subsec:problem_formulation}) and then detail our meta-trained BERT with  in-context tuning approach for reading comprehension AS (Section~\ref{subsec:model}).

% SUBSECTION %
\subsection{Problem Formulation}
\label{subsec:problem_formulation}

\begin{table}
\caption{Text snippets from an example grade 8 reading comprehension item.}
\label{tab:example_task}
\centering
\begin{tabular}{p{0.14\textwidth} p{0.69\textwidth}p{0.14\textwidth}}

\toprule
\textbf{Passage} & Long ago, a poor country boy left home to seek his fortune. Day and night he traveled, stopping to eat at inns along \ldots \\
\textbf{Question} & Describe what kind of person the merchant is. Give one detail from the story to support your answer.\\

\midrule

\multirow{4}{*}{\textbf{Dataset}} & \multicolumn{1}{c}{\textbf{Student Responses}} & \multicolumn{1}{c}{\textbf{Scores}}\\
\cmidrule{2-3}
& ``The merchant is a very optimistic and persevering person. For example, it states how he kept walking on even with an empty stomach. This shows that he is hopeful and not willing to give up after he had come so far." & \multicolumn{1}{c}{3} \\
& ``the merchant is a determined and honest person" & \multicolumn{1}{c}{2} \\
& ``Dishonest because he didn't want to pay for the eggs." & \multicolumn{1}{c}{1} \\
\bottomrule

\end{tabular}
\end{table}

The NAEP AS challenge features $20$ reading comprehension items from grade 4 and grade 8.
Each reading comprehension item, indexed by $i$,  
is associated with 1) a long reading passage text $P_{i}$, 2) a short question text $Q_{i}$, 3) a scoring rubric, 
and 4) a large training dataset of human scored  
student responses $D_{i}^{train} = \{(x_{j}, y_{j})\}$. 
The goal is for trained models to predict human scores for student responses in the test dataset $D_{i}^{test} = \{(x_{j}^{target})\}$ for all tasks $i$. A key observation is that pairs of items $(l, m)$, one from grade 4 ($l$) and one from grade 8 ($m$), 
often share passage $(P_{l}=P_{m})$ and question $(Q_{l}=Q_{m})$ texts. This linkage across items motivates our approach of leveraging shared item semantics by not treating items independently. Instead of training a separate model for each item, as in existing approaches, we meta-train a single shared AS model to perform in-context learning across all items. 
A snapshot of an example item from the NAEP question tool\footnote{\url{https://nces.ed.gov/NationsReportCard/nqt/Search}} is shown in Table~\ref{tab:example_task}.

% SUBSECTION %
\subsection{Our Approach: Meta-trained BERT with In-context Tuning}
\label{subsec:model}

\subsubsection{LM Fine-tuning Setup.}
We first describe our LM fine-tuning approach for reading comprehension AS. 
We use a pre-trained BERT~\cite{devlin2018bert} model as the base LM for AS.\footnote{We also tried GPT-2~\cite{radford2019gpt2} and reached similar scoring performance. Therefore, for simplicity of exposition, we only detail our approach with BERT as the base LM.} 
We use a combination of the passage $P_{i}$, question $Q_{i}$, and student response text $x_{j}^{target}$ as input to the BERT model. We add $\texttt{[SEP]}$, the separator token in BERT's vocabulary, between these parts to help the model differentiate input text segments with different purposes.
Since the average passage length was $657$ words, directly inputting every token in the passage text exceeds the maximum input length of $512$ tokens for BERT. 
As a workaround, we use a separate, fixed BERT model (we do not fine-tune its parameters) to encode the passage text and use its $\texttt{[CLS]}$ embedding vector as input to the BERT model for AS. We experiment with two ways to encode the passage text: either the entire passage as one token, or each sentence as one token. 
For the downstream AS task, we feed the $\texttt{[CLS]}$ embedding vector of the BERT AS model to a linear classification layer followed by softmax~\cite{dlbook} during fine-tuning.

\subsubsection{Meta-training BERT with In-context Learning.}

\begin{figure}[tp]
    \centering
    \includegraphics[width=0.95\linewidth]{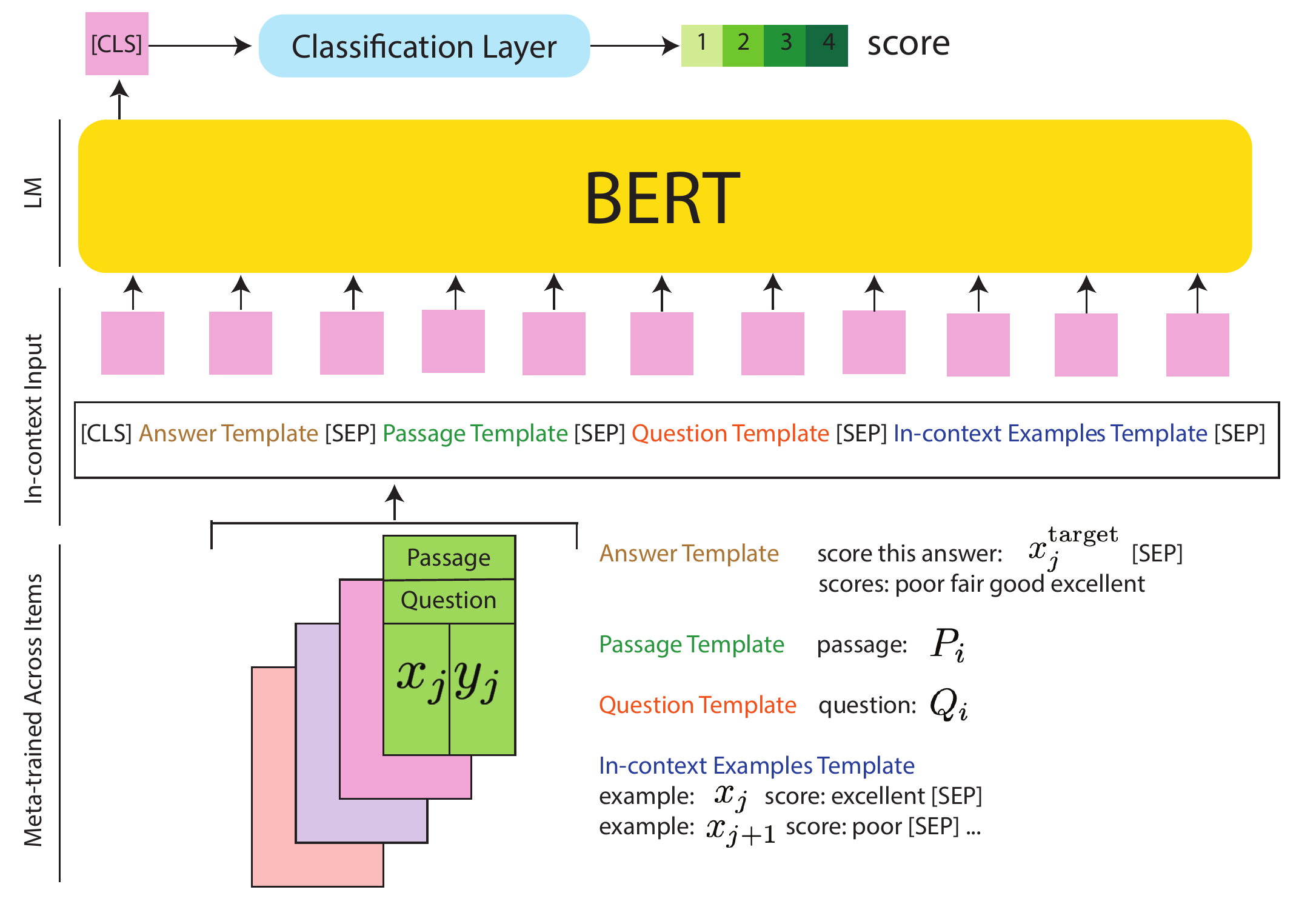}
    \caption{Meta-trained BERT via in-context tuning. Best viewed in color.}
    \label{fig:meta_bert_model}
\end{figure}

We now detail the key component of our approach, in-context learning of a shared BERT AS model across all items. 
Our approach is to use a carefully-designed input structure to provide context to the shared BERT AS model and associate it with each specific item. 
To achieve this, following recent approaches for meta-training-based in-context learning~\cite{chen2021meta,min2021metaicl}, we add \textit{in-context examples} which are (response, score) pairs from the training set, to the input. 
Intuitively, these examples provide further context about each item and enable the BERT AS model to focus on learning to associate responses to scores given the passage and the question, thus enabling knowledge sharing across items. 

Figure~\ref{fig:meta_bert_model} shows the input structure to the BERT AS model. 
We build the model input by concatenating the target student response text $x_{j}^{target}$ that needs to be scored, passage $P_{i}$, question $Q_{i}$, and $K$ response-score examples $E_{i} \subseteq D_{i}^{train}, |E_{i}|=K$. 
Moreover, we convert the numeric scores $y_j$ in the in-context examples $E_{i}$ to meaningful words 
as: $\{1:\text{poor}, 2:\text{fair}, 3:\text{good}, 4:\text{excellent}\}$ before adding them to the model input.
We add semantically meaningful task instructions to each input component as shown in Figure~\ref{fig:meta_bert_model}. 

Since each item $i$ has a different range of possible scores $y_j \in [s_{min}=\{1\}, s_{max} = \{2,3,4\}]$, we add possible score classes for $x_{j}^{target}$ as explicit options~\cite{wei2021finetuned} in the input to make the model aware of valid scores as seen in the answer template in Figure~\ref{fig:meta_bert_model}. Furthermore, we mask out the invalid score classes for each response before computing the softmax. This masking procedure ensures that training loss is backpropagated over valid score classes only, enabling us to
meta-train a single shared model for all items.

\subsubsection{Meta-training.}
We train our AS model on the union of training datasets for all items $\bigcup_{i=1}^{20} D_{i}^{train}$. 
Let $\theta$ represent the model parameters to be learned, i.e., BERT parameters which are fine-tuned after initialization from their pre-trained values and the classification layer parameters which are learned from scratch. 
The in-context learning objective $\mathcal{L}_i$ for an item is simply the cross entropy, i.e., negative log-likelihood loss~\cite{dlbook}:
$$ \mathcal{L}_{i}(\theta) = \sum_{(x_{j}^{target}, y_{j}^{target}) \in D_{i}^{train}} [-\log p_\theta (y_{j}^{target} | x_{j}^{target}, P_{i}, Q_{i}, E_{i}) ], $$
while the objective across all $20$ items is given by:
$$ \mathcal{L}(\theta) = \sum_{i=1}^{20} \mathcal{L}_{i}(\theta). $$
In contrast, existing approaches result in a separate set of model parameter $\theta_i$ for each item, which increases the number of model parameters and thus storage space requirement by $20\times$.

%
% SECTION %
%
\section{Experimental Evaluation}

In this section, we detail our experiments on the training dataset provided by the NAEP AS challenge. We conduct both quantitative evaluations on scoring accuracy of our approach and qualitative evaluations on the error types and fairness aspects of our approach.

% SUBSECTION %
\subsection{Dataset Details}

Each item in the NAEP AS Challenge has responses from either one of two formats: short (around one sentence) or extended (up to around five sentences) responses, leading to different score classes. The minimum score for all items is $1$, with the maximum being either $2$ or $3$ for short responses and $3$ or $4$ for extended responses. Passages and questions have around $657$ and $27$ words on average, respectively. 
Our training dataset (combined from the challenge's training and validation datasets) contains on average more than $18,000$ human-scored student responses per item. 
Each response in the training dataset is scored by either one or two human raters. 
We are also provided with the gender and ethnicity of the student who submitted each response. 
The average student response length is $19$ and $37$ words for grades $4$ and $8$, respectively. 
There are $4$ unique passages and $12$ unique questions; $8$ pairs of items (one from grade $4$ and one from grade $8$) share the same passage and question text. The $4$ remaining items have unique questions.

% SUBSECTION %
\subsection{Metrics and Baselines}
\label{subsec:baselines}

Following the NAEP AS challenge, we use quadratic weighted Kappa (QWK), a popular metric for AES, as the accuracy metric. QWK examines the agreement between the predicted scores and the ground truth human scores for ordered score classes. 
We compare our approach, dubbed \textbf{meta-trained BERT with in-context} tuning, to various baselines including existing approaches for AES and several variants of our approach, listed below. The latter serves as an ablation study to verify the importance of each component of our approach. 
\begin{itemize}
    \item \textbf{Human} provides an estimated upper bound on scoring accuracy given by the inter-rater agreement. For each item, we calculate the QWK on around $550$ training responses that are scored by two human raters. 
    \item \textbf{Majority} simply scores each response by the most frequent score for that item in the training dataset.
    \item \textbf{Feature Engineering} uses a feature-engineering approach with a selected set of features listed in Table~\ref{tab:features} for a random forest classifier.
    \item \textbf{Stacked LSTM} uses a stack of two LSTMs on Word2Vec embeddings followed by a non-linear output layer using ReLU activation.
    \item \textbf{Clusering + Classification}, uses cluster indicators of student responses (computed via K-means clustering on their Word2Vec embeddings) in conjunction with other features for a random forest classifier. 
    \item \textbf{BERT (response)} fine-tunes a BERT model for each item independently using only response text as input, resulting in $20$ different AS models.
    \item \textbf{BERT (passage+question+response)} adds passage and question information as input to the setup of BERT (response). 
    \item \textbf{BERT in-context} adds in-context examples as input to the setup of BERT 
    (passage+question+response)
    providing further context on each item.
    \item \textbf{BERT multi-task} uses a multi-task learning approach and fine-tunes a single shared BERT AS model with $20$ separate classification layers, one for each item. We backpropagate the loss for a student response from item $i$ to train only the corresponding classification layer while fine-tuning the shared BERT model on all items.
\end{itemize}

\begin{table}[tp]
\caption{\label{tab:features} Human-designed features used in baseline models.}
\centering
\begin{tabular}{p{0.14\linewidth} p{0.83\linewidth}}

\toprule
Type & Features \\
\midrule
Length & Counts of words, characters, stop words, and punctuation. Character to word ratio. \\
Syntax & Counts of lemmas, nouns, verbs, adjectives, adverbs, and conjunctives.\\
Readability & Automated Readability Index, Coleman Liau Index, Dale Chall Score, Smog Index, and Flesch Score.\\
\bottomrule
\end{tabular}
\end{table}

% SUBSECTION %
\subsection{Implementation Details}

\subsubsection{Preprocessing: Spell Check.}
We automatically correct spelling errors
in student responses for two reasons: 1) the scoring rubrics have clear instructions on ignoring such errors and 2) to transform the response text written by students to be similar to what BERT uses in pre-training. Spelling errors are frequently seen in grade $4$ responses but are less frequent in grade $8$.
We experiment with training models on student responses with and without spell checking; The spell checked version leads to an improvement of around $0.5\%$ on QWK. 
We use NeuSpell's BERT based checker~\cite{jayanthi-etal-2020-neuspell} to correct spelling errors 
on all student responses.

\subsubsection{Experimental Setup.}
For our local evaluation, we perform five-fold cross-validation on the training dataset for each item and use three folds for training, one fold for validation to perform early stopping, and one fold for testing. 
For double-scored student responses where the human-rater scores do not agree, we use the score by the first rater, following the challenge specification. 

\begin{table}[tp]
\caption{\label{tab:results} Experimental results averaged across reading comprehension items in the NAEP AS challenge dataset. Model performance in Quadratic Weighted Kappa (higher is better). \textbf{Bold} indicates best result except human performance.}
\centering
\begin{tabular}{p{0.59\linewidth}p{0.18\linewidth}p{0.18\linewidth}}

\toprule
Approach & Avg. QWK & \textit{p}-value \\
\midrule
Human & $0.878$ & -\\
Majority & $0.527$ & -\\
Feature Engineering + Random Forest & $0.443$ & -\\
Stacked LSTM & $0.657$ & -\\
Clustering + Classification & $0.709$ & -\\
BERT (response) & $0.828$ & -\\
BERT (passage+question+response) & $0.828$ & $0.414$\\
BERT in-context & $0.833$ & $0.001$\\
BERT multi-task & $0.833$ & \num{1.6e-4}\\
Meta-trained BERT in-context & $\mathbf{0.841}$ & \num{9.6e-5}\\
\bottomrule

\end{tabular}
\end{table}

We use a pre-trained BERT~\cite{devlin2018bert} model with $110M$ parameters; all implementation was done using the HuggingFace~\cite{wolf-etal-2020-transformers} transformers library. 
We use the Adam optimizer, a batch size of $32$, a learning rate of $2 \cdot 10^{-5}$, and a maximum input length of $512$ tokens.
We do not perform any hyperparameter tuning and simply use the default parameters for BERT.
We fine-tune all BERT-based models for $10$ epochs since they reach optimal performance on the validation set by around $5$ epochs. 
Our meta-trained BERT with in-context tuning takes $6$ hours per training epoch on a single NVIDIA RTX $8000$ GPU.

For our meta-trained BERT model with in-context tuning, for each training response, we randomly sample up to $25$ in-context examples per score class from the training dataset of the corresponding item. 
We ensure at least one example from each valid scoring class so that the set of examples cover all classes and is diverse. 
Given the restriction on input length for BERT, we truncate each example to $70$ tokens. 
At test time, for a target student response to be scored, we repeat the aforementioned process of randomly sampling examples for $8$ times and average the predicted score class probabilities.

\begin{table}[tp]
\caption{\label{tab:results_shared} Results averaged over shared vs. non-shared reading comprehension items from grade 4 (G4) and grade 8 (G8). Model performance in Quadratic Weighted Kappa (higher is better). \textbf{Bold} indicates best result except human performance.}
\centering
\begin{tabular}{p{0.44\linewidth}p{0.14\linewidth}p{0.11\linewidth}p{0.14\linewidth}p{0.11\linewidth}}

\toprule
Approach & Non-shared G4 & Shared G4 & Non-shared G8 & Shared G8 \\
\midrule
Human & $0.860$ & $0.899$ & $0.809$ & $0.885$\\
BERT (response) & $0.825$ & $0.842$ & $0.763$ & $0.840$ \\
BERT (passage+question+response) & $0.820$ & $0.843$ & $0.762$ & $0.839$ \\
BERT in-context & $0.822$ & $0.847$ & $\mathbf{0.771}$ & $0.843$ \\
BERT multi-task & $0.817$ & $0.848$ & $0.765$ & $0.845$ \\
Meta-trained BERT in-context & $\mathbf{0.826}$ & $\mathbf{0.856}$ & $\mathbf{0.771}$ & $\mathbf{0.853}$ \\
\bottomrule

\end{tabular}
\end{table}

% SUBSECTION %
\subsection{Results and Analysis}
We report the average QWK across all items and all cross validation folds for all approaches in Table~\ref{tab:results}. LM-based approaches consistently outperform non-LM-based approaches, by a significant margin, while our approach, meta-trained BERT in-context, performs best. We perform paired t-tests among LM-based approaches using BERT (response) as the baseline; We see that learning a single shared model across items achieves statistically significant improvement over learning one model per item. This observation validates our intuition that we can leverage shared information across reading comprehension items to improve AS accuracy. 
We also observe that, surprisingly, adding passage and question text as input in addition to response text does not lead to improved performance. This observation suggests that our models are trained for AS but cannot perform reading comprehension; We discuss this in detail in Section~\ref{sec:conc}. 
Table~\ref{tab:results_shared} lists performance of LM-based approaches on shared vs. non-shared items. Overall, our approach performs better than BERT (response), more on items that are shared across grades $4$ and $8$ and less on items that are not shared. This observation suggests that our approach may struggle to generalize to previously unseen items, which is an important avenue for future work. 
For most items, some LM-based approach reach a level of AS accuracy close to that for humans (as measured in QWK between two human scorers). While there is still room for improvement, it is important to gauge the limit of AS approaches due to intrinsic noise in data.

\subsubsection{Qualitative Error Analysis.} We randomly sample $100$ responses to the item ``describe what kind of person the merchant is \ldots'' as seen in Table~\ref{tab:example_task} that our approach scores incorrectly.
We identify $5$ main error types and list them in Table~\ref{tab:error_analysis} with made-up examples that reflect our observation: 1) spelling and grammar errors (not caught by spell check), 2) human error and subjectivity, 3) infrequent correct responses (our approach struggles at recognizing correct responses that do not occur frequently), 4) imitation (incorrect responses with content/structure that mimic correct responses), 5) character coreference (referring to a character different than the one that the item asks about). These observations reveal further ways to improve LM-based AS approaches for reading comprehension. 

\begin{table}[tp]
\caption{\label{tab:error_analysis} Illustration of responses that our approach tends to score incorrectly.}
\centering
\setlength{\tabcolsep}{0.4em}
\begin{tabular}{p{0.2\linewidth}p{0.48\linewidth}m{0.1\linewidth}m{0.1\linewidth}}

\toprule
\centering Error Type & \centering Example Student Response & \centering Predicted Score & Ground Truth \\
\midrule

Spelling/grammar & ``\textit{mearchant are} a good man because he \textit{played} the innkeeper and kept his \ldots" & \multicolumn{1}{c}{2} & \multicolumn{1}{c}{3} \\

Human error \& subjectivity & ``Long ago a poor country boy left home to seek his fortune. Day and night he \ldots" & \multicolumn{1}{c}{1} & \multicolumn{1}{c}{3} \\

Infrequent correct answers & ``merchant is described as \textit{brave} as he got on a ship and visited multiple ports \ldots" & \multicolumn{1}{c}{2} & \multicolumn{1}{c}{4} \\

Imitation & ``The merchant is dishonest because he \textit{doesn't want to pay} for the eggs \ldots" & \multicolumn{1}{c}{3} & \multicolumn{1}{c}{1} \\

Character coreference & ``The \textit{merchant} is greedy because he gives the \textit{innkeeper} eggs but when the boy \ldots" & \multicolumn{1}{c}{4} & \multicolumn{1}{c}{1} \\

\bottomrule

\end{tabular}
\end{table}

\subsubsection{Fairness Studies.} 
\begin{figure}
\vspace{-.2cm}
    \centering
    \includegraphics[width=0.49\linewidth]{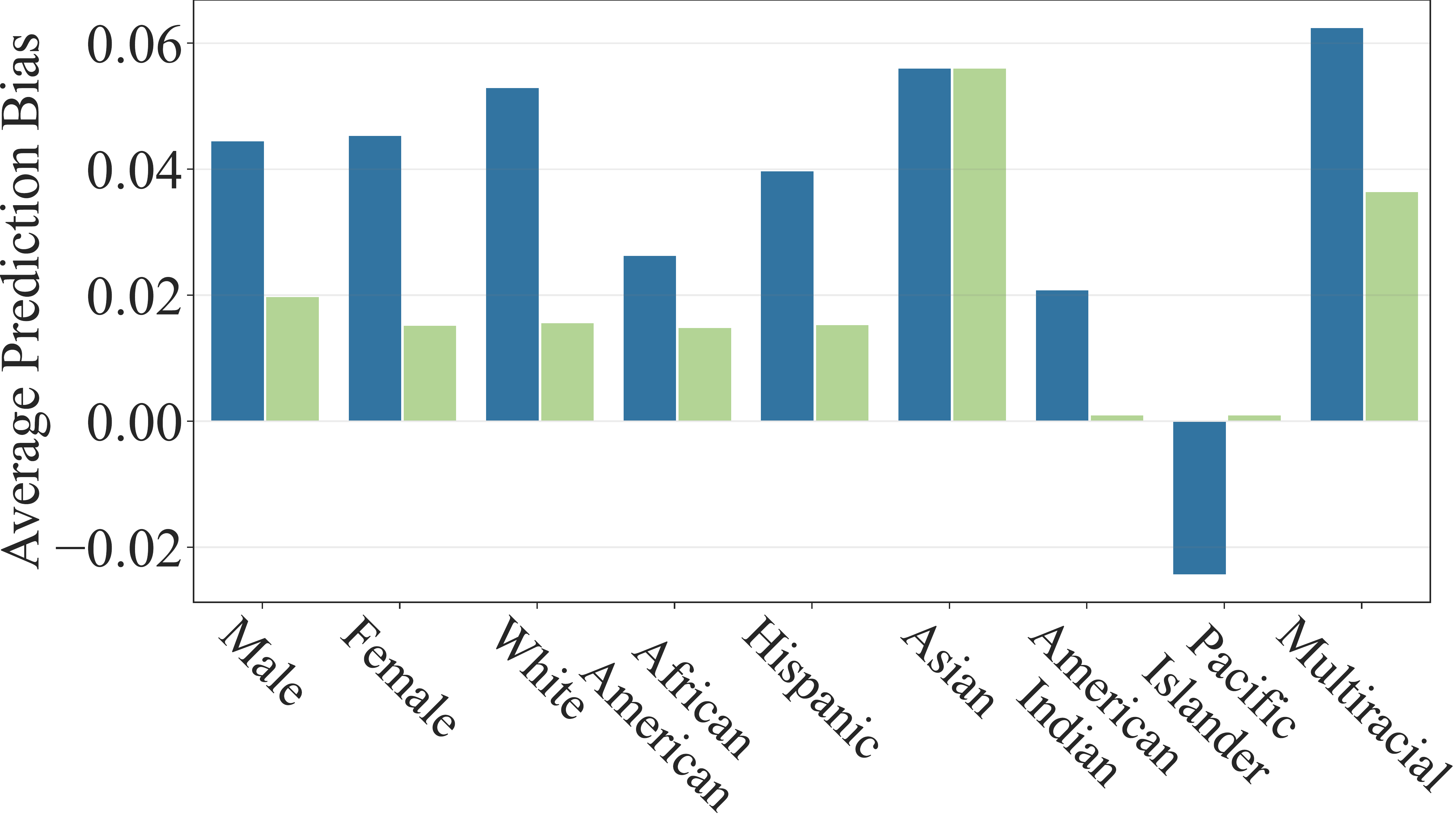}
    \hspace{1pt}
    \includegraphics[width=0.46\linewidth]{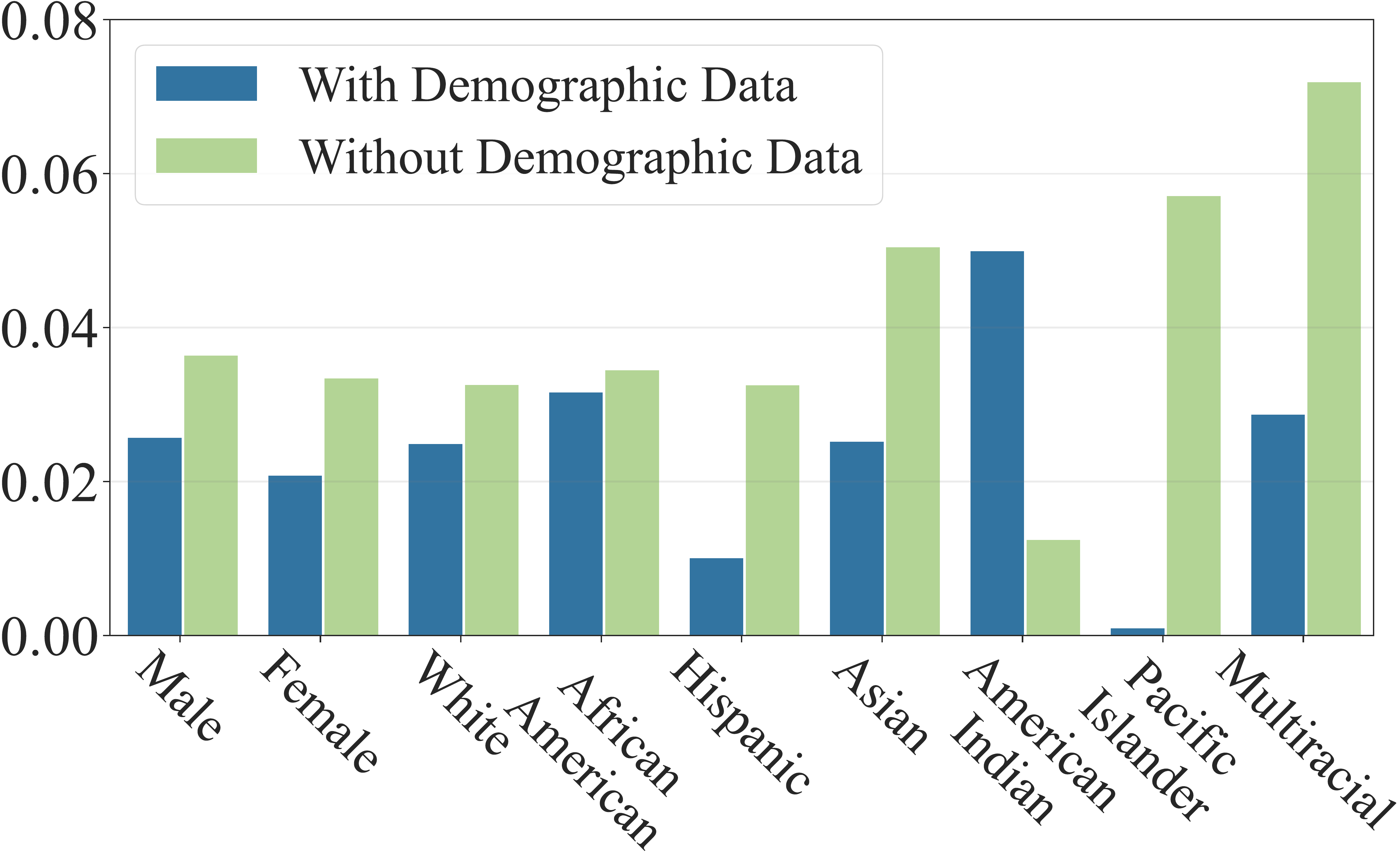}
    \vspace{-.1cm}
    \caption{Prediction biases grouped by student demographics with and without demographic data as input for a shared item used in both grade 4 (left) and grade 8 (right).}
    \label{fig:fairness}
    \vspace{-.2cm}
\end{figure}

We conduct a preliminary investigation into the fairness aspects of our approach using the demographic information in the NAEP AS challenge dataset. There are a total of 9 demographic groups, each corresponding to a gender or an ethnicity group. For students in each group, we compute the average prediction bias, i.e., the average difference between the predicted score and the actual score. A positive bias means that the model overestimates the score for the group while a negative bias means that the model underestimates the score. 
We experiment with two settings: not using the demographic information (gender/ethnicity) as input vs.\ using it as input to the BERT model. We test both settings since it is not clear whether leaving out demographic information or other sensitive attributes results in more fair models~\cite{yu2021should}. 
When using demographic information as input, we prepend the input with textual instructions such as ``score this answer written by a Female Asian student'', according to the demographics of the student.
Figure~\ref{fig:fairness} shows the prediction biases for a shared item in both grades 4 and 8. We see that our approach tends to overestimate the score, except for Pacific Islanders/American Indians on grade 4, which is possibly due to the small number of such students in the training data. 
In general, the biases are within an acceptable range (the same conclusion is also reached on the official challenge test set). 

%
% SECTION %
%
\section{Discussion, Conclusions, and Future Work}
\label{sec:conc}
In this paper, we detailed our solution to the NAEP automated scoring challenge, an in-context BERT fine-tuning approach that is suitable for reading comprehension items. We now discuss some important observations and outline several avenues for future work towards the scalable deployment of language model-based automated scoring approaches in real-world scenarios. 

First and foremost, we observe that our best-performing model takes as input only the in-context examples, question text and the target response but not the reading comprehension passage. Although this approach is due to the technical limitation that these passages are often longer than what language models can input, this observation means that our model does not master the task of reading comprehension; instead, it only learns to assess how appropriate the response is given the item. Future work should find ways to efficiently pass information contained in the passage to the model using a small number of tokens, such as retrieving only relevant sentences \cite{drqa}. Models that learn to score responses in the context of the passage will be better suited for generalization to new items. 

Second, we observe that adding an additional spell checking step improved automated scoring accuracy and that using BERT only as a text featurization tool without fine-tuning leads to a significant drop in scoring accuracy. These observation suggest that BERT is not adept at student-generated text. Future work should pre-train base language models on student-generated text before applying them to downstream tasks.

Third, we observe that our models do not offer any explanation on their score predictions due to the black-box nature of language models. Future work should incorporate grading rubrics into the scoring model or use other ways to improve model interpretability, such as using attention maps \cite{attnmap}. 

Fourth, we observe that our models, without explicitly controlling for fairness, exhibit unavoidable but modest biases toward students in different demographic groups. Future work should incorporate fairness regularization into the training objective to promote fairness across students \cite{Zafar:Fairness:2017}. 

\subsubsection{Acknowledgements.}
Fernandez, Ghosh, and Lan are partially supported by the National Science Foundation under grants IIS-1917713 and IIS-2118706.


\bibliographystyle{splncs04}
\bibliography{references}

\end{document}